\documentclass[preprint,12pt]{elsarticle}
\usepackage{amssymb}
\usepackage{longtable,booktabs,array}
\usepackage{multirow}
\usepackage{calc} % for calculating minipage widths
\usepackage{url}

\usepackage{rotating}

\journal{Applied Ergonomy}

\begin{document}

\begin{frontmatter}

%% Title, authors and addresses

%% use the tnoteref command within \title for footnotes;
%% use the tnotetext command for theassociated footnote;
%% use the fnref command within \author or \address for footnotes;
%% use the fntext command for theassociated footnote;
%% use the corref command within \author for corresponding author footnotes;
%% use the cortext command for theassociated footnote;
%% use the ead command for the email address,
%% and the form \ead[url] for the home page:
%% \title{Title\tnoteref{label1}}
%% \tnotetext[label1]{}
%% \author{Name\corref{cor1}\fnref{label2}}
%% \ead{email address}
%% \ead[url]{home page}
%% \fntext[label2]{}
%% \cortext[cor1]{}
%% \affiliation{organization={},
%%             addressline={},
%%             city={},
%%             postcode={},
%%             state={},
%%             country={}}
%% \fntext[label3]{}

\title{Human-Cobot collaboration's impact on success, time completion, errors, workload, gestures and acceptability during an assembly task}

%% use optional labels to link authors explicitly to addresses:
%% \author[label1,label2]{}
%% \affiliation[label1]{organization={},
%%             addressline={},
%%             city={},
%%             postcode={},
%%             state={},
%%             country={}}
%%
%% \affiliation[label2]{organization={},
%%             addressline={},
%%             city={},
%%             postcode={},
%%             state={},
%%             country={}}

\author[inst1]{Étienne Fournier}
\affiliation[inst1]{organization={Univ. Grenoble Alpes \--- LIP/PC2S},%Department and Organization
            city={Grenonle},
            postcode={38000},
            country={France}}

\author[inst1]{Christine Jeoffrion}
\author[inst2]{Belal Hmedan}

\affiliation[inst2]{organization={Univ. Grenoble Alpes \---LIG},%Department and Organization
            city={Grenonle},
            postcode={38000},
            country={France}}

\author[inst2]{Damien Pellier}
\author[inst2]{Humbert Fiorino}
\author[inst2]{Aurélie Landry}

\begin{abstract}
The 5.0 industry promotes collaborative robots
(cobots). This research studies the impacts of cobot collaboration using an experimental setup. 120 participants realized a simple and a complex assembly task. 50\% collaborated with another human (H/H) and 50\% with a cobot (H/C). The workload and the acceptability of the cobotic collaboration were measured. Working with a cobot decreases the effect of the task complexity on the human workload and on the output quality.
However, it increases the time completion and the number of gestures (while decreasing their frequency). The H/C couples have a higher chance of success but they take more time and more gestures to realize the task. The results of this research could help developers and stakeholders to understand the impacts of implementing a cobot in production chains.
\end{abstract}


\end{frontmatter}

\section{Introduction}

Since 2010, the industry 4.0 pushes the implementation of intelligent
technologies in the industrial workplace to increase productivity
(Oztemel \& Gursev, 2020). In 2020, the industry 5.0 was launched by the
European Commission (Xu et al., 2021). This new industry calls on
industry players to put humans back at the heart of the industrial
system. It promotes Human/Cobot (H/C) collaboration. Therefore, cobots
are being increasingly implemented in the industry. ``The cobot market
is rapidly expanding, and the academic literature is similarly growing''
(Knudsen \& KaiVo-Oja, 2020, p. 13). The term cobot refers to "a robotic
device which manipulates objects in collaboration with a human operator"
(Colgate et al., 1996, p. 1). Their use could be a way of reducing
constraints and increasing performances without replacing the human
factor (Peshkin \& Colgate, 1999). Cobots work closely to humans;
therefore, they are equipped with security features to prevent injuries
(Djuric et al., 2019; Gualtieri et al., 2021). However, implementing a
cobotic system in the workplace can have other negative effect. Authors
have identified a positive effect of the H/C collaboration, such as
reduction of musculoskeletal disorders risks factors (Bouillet et al.,
2023; Schoose, 2022). However, this implementation puts other strains on
the operator and can decrease the output quality (Schoose, 2022).
Therefore, it seems important to test the effect of the H/C
collaboration on human operators (Kildal et al., 2018). We choose to
focus on two types of reports of HC collaboration that are particularly
interesting to us because they test real-life direct interaction with
cobots: feedbacks from the field which focus on the operators and their
perceptions (Barcellini et al., 2023; Catchpole et al., 2019; Cheon et
al., 2022; Pauliková et al., 2021; Schoose, 2022; Schoose et al., 2023)
and feedbacks from experiments studying their impacts in laboratories'
conditions (Alarcon et al., 2021; Fournier et al., 2022; Hopko et al.,
2023; Varrecchia et al., 2023). The cobot adaptation capacities could
help dealing with variabilities in the workplace (Hiatt et al., 2011).
These studies revealed that H/C collaboration improved worker's
coordination, decreased physical effort (Varrecchia et al., 2023),
improved efficiency (Fournier et al., 2022) and decreased exposure to
risks (Pauliková et al., 2021). However, studies have reported negative
effects of implementing cobotic systems (Cheon et al., 2022; Fournier et
al., 2022; Schoose, 2022; Schoose et al., 2023). There have been cases
of loss of job identity (Cheon et al., 2022), increased task's time
completion (Fournier et al., 2022; Mariscal et al., 2023) and lack of
cobot adaptability to human constraints (Schoose, 2022; Schoose et al.,
2023). To be implemented in the industry, cobotic collaboration needs to
be effective in terms of results (higher quality of output, faster
completion time, lower number of gestures and lower number of errors).
Also, realizing the task with less gestures and with a decreased mental
workload from the human could be interesting for the industry as cobotic
collaboration is designed to increase the operator's well-being (Bogue,
2016; European Commission, 2021; Salunkhe et al., 2019). The mental
workload can be representative of the operator's wellbeing at work
(Fruggiero et al., 2020), therefore, its study in the context of cobotic
collaboration is important. Those variables are all connected to the
concept of usability (Bevan et al., 2015) which is ``the extent to which
a product can be used by specified users to accomplish specific goals
with effectiveness, efficiency and satisfaction in a particular context
of use'' (ISO, 2018). The effectiveness (degree of completion and
accuracy) of the cobotic collaboration can be measured by the number of
errors and the success of the task. Efficiency (minimal of resources
used) can be measured with the time completion and the number of
gestures and workload. Finally, satisfaction (meet the user's
expectations) can be measured with the acceptability. According to
Fournier et al. (2022), a cobotic collaboration could be useful in the
industry to improve the success of a task, unlike Schoose et al. (2023)
who reported a less qualitative output. Also, some studies demonstrate a
lower time completion of the task (Fager et al., 2019) and others report
the opposite (Fournier et al., 2022; Mariscal et al., 2023). Concerning
the number of errors, some authors demonstrated a decrease with a
cobotic collaboration (Fournier et al., 2022), while other showed an
increase (Mariscal et al., 2023). The cobotic collaboration's impact on
workload is also unclear, a study found a decrease working with cobots
(Kildal et al., 2019), while another found no significative change
(Fournier et al., 2022). Furthermore, only one study (which we
conducted) showed a decrease in the number of gestures with a cobotic
collaboration (Fournier et al., 2022). Those different variables require
further investigations before allowing cobots to be massively
implemented in the industry in order to guarantee their acceptation
(Bobillier Chaumon, 2016). If they are not accepted by the operators,
they won't use it or they won't use well (Cippelletti, 2017). Nielsen
developed a model of acceptability (Nielsen, 1994) that included the
concept of ease of use of the system and usefulness (ease for the system
to achieve the set goal). These concepts were used in the Unified Theory
of Acceptance and Use of Technology (UTAUT), a model to assess the
acceptation of a technology (Venkatesh et al., 2003). In this model, the
authors investigate four factors that have positive impacts on the use
intention. These factors include:

\begin{itemize}
\item
  Expected performance (or perceived usefulness): the degree to which
  the individual believes using the system will help him achieve his
  goal (Venkatesh et al., 2003).
\item
  Expected efforts (or perceived ease of use): the degree to which the
  user perceives the system easy to be used (Venkatesh et al., 2003).
\item
  Social influence: the degree to which the individual believes that
  people who are important to him believe that he should use the system
  (Venkatesh et al., 2003).
\item
  Facilitating conditions: the degree to which an individual perceives
  that the characteristics of his environment will help to use the
  system (Venkatesh et al., 2003).
\end{itemize}

These different factors predicted 56\% of behavioral intention and 40\%
of use of mobile applications on smartphones by 1,512 participants
(Venkatesh, 2012). The authors of the UTAUT completed their original
model analyzing over 500 articles dealing with technology adoption,
adding 3 independent variables (Venkatesh, 2007, in Martin, 2018). The
hedonic motivation is the degree of perceived pleasure associated with
using the technology. The monetary cost is the perceived financial cost
of using the technology. And the ``user habits'' is the individual habit
of using technology. UTAUT2 predict 74\% of behavioral intention and
52\% of usage (Venkatesh, 2012).

The goal of this study is to determine if the human's success in a task
(Fournier et al., 2022; Schoose et al., 2023), the time completion
(Fager et al., 2019; Fournier et al., 2022; Mariscal et al., 2023), the
number of errors (Fournier et al., 2022; Mariscal et al., 2023), the
workload (Fournier et al., 2022; Kildal et al., 2019), the number of
gestures (Fournier et al., 2022) and acceptability of the collaboration
are different whether the collaboration is Human/Human (H/H) or H/C by
controlling certain variabilities. Variabilities are the variations to
which each work situation is subject to (Brangier \& Valléry, 2021).
They may be individual, i.e. originating from the operator himself, or
external (Guérin et al., 1997). For example, a cobotic collaboration is
more efficient than its absence when realizing an easy task (Fournier et
al., 2022), but it has yet to prove itself flexible enough to be
efficient when realizing a complex one, closer to a real-life work
situation (Kildal et al., 2018). This task demand (or complexity)
variability has to be investigated. Also, to guide the cobot conception,
it was necessary to determine if the arm use of the cobot could have a
different impact whether it was in tune with the dominant hand of the
participant (e.g. using the right arm facing the participant when he was
right-handed) or not (e.g. using the right arm facing the participant
when he was left-handed). For those reasons, in our research, we aimed
to determine how the cobot adapts to those two variabilities (task
demand and dominant hand) during an assembly task. We too decided to
create a questionnaire to assess acceptability of the cobotic
collaboration. In fact, no research, as far as we know, has investigated
the acceptability of cobots before and after use, this is one objective
of the study presented. We also want to measure the impact of H/C
collaboration on the human workload, which is the amount of mental
resources used for a specific task (Verhulst, 2018), and the number of
gestures (Catchpole et al., 2019). We define collaboration as two agents
working on the same task in the same time (Matheson et al., 2019; Müller
et al., 2017).\\

\noindent Our operational hypotheses are:

\textbf{H1}: The increase of the \textbf{task demand} has a negative
effect on the participants' workload, success, errors, time completion
and gestures. This effect is lower in the H/C condition than in the H/H
condition.

\textbf{H2}: Adapting the cobot to the operator\textquotesingle s
\textbf{dominant hand} has a positive effect on the
participants\textquotesingle{} workload, success, errors, time
completion and gestures.

\textbf{H3}: H/C \textbf{collaboration} has more positive effects on the
success rate, the number of errors, the completion time, the workload
and the number of gestures compared to the H/H collaboration.

\textbf{H4}: Participant's \textbf{acceptability} \textbf{score} of the
H/C collaboration is higher in the H/C condition than in H/H condition.

\section{Material and methods}

This study was pre-registered on OSF and registered by the laboratory's
director under the authority of the university.

\subsection{Participants}

120 subjects participated in the study (90 female, 29 males and 1
non-binary). 81\% of the participants were right-handed, 16\%
left-handed and 3\% fluid with both hands. This is still higher than in
the general French population that counts 12,9\% of left-handed
(Lesgauchers, 2017). Only 8\% of the participants declared having worked
with a robot before. Participants were adults, without any color vision
impairment and with no pain in the upper limbs prior to the experiment.
They were recruited by a post on the website of the university.
Psychology students were awarded point in certain courses. Participants
were divided into two different groups.

\subsection{Procedure}

\begin{figure}
\centering \includegraphics[width=5.6406in,height=4.33988in]{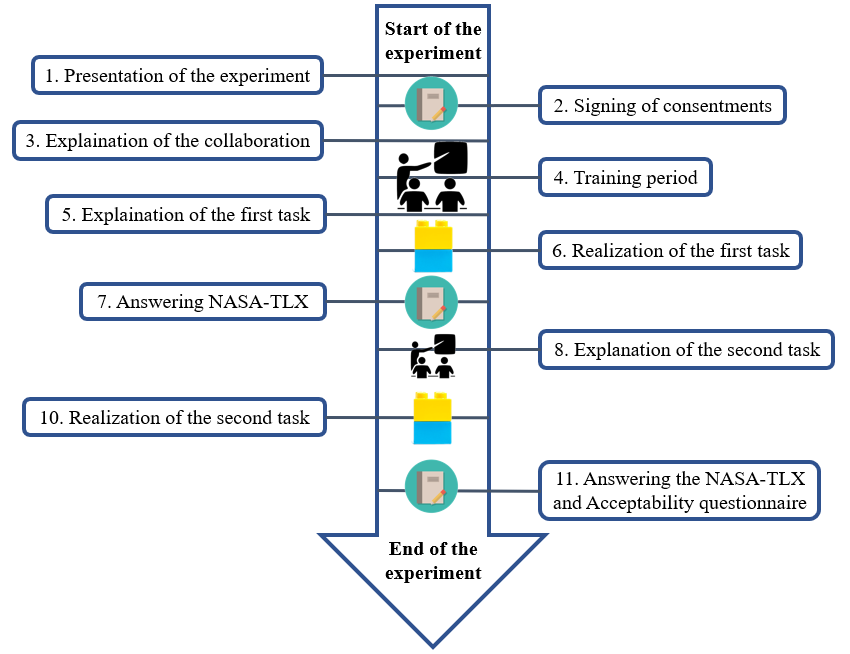}

\caption{Illustration of the experimental procedure}
\end{figure}

The experimental procedure is illustrated Figure 1.
 In our study, two conditions were set up, participants were randomly
assigned to two different groups corresponding to those conditions. The
use of an intergroup rather than an intragroup modality was justified by
logistical constraints and by the desire to avoid learning and fatigue
effects. The first group (N=61) realized two Duplo assembly tasks
(simple and complex in a randomized order) during a H/H collaboration.
The second group (N=59) realized the same tasks with a cobot (H/C
condition; cf. Figure 2).

\begin{figure}
\centering\includegraphics[width=4.35317in,height=3.47639in]{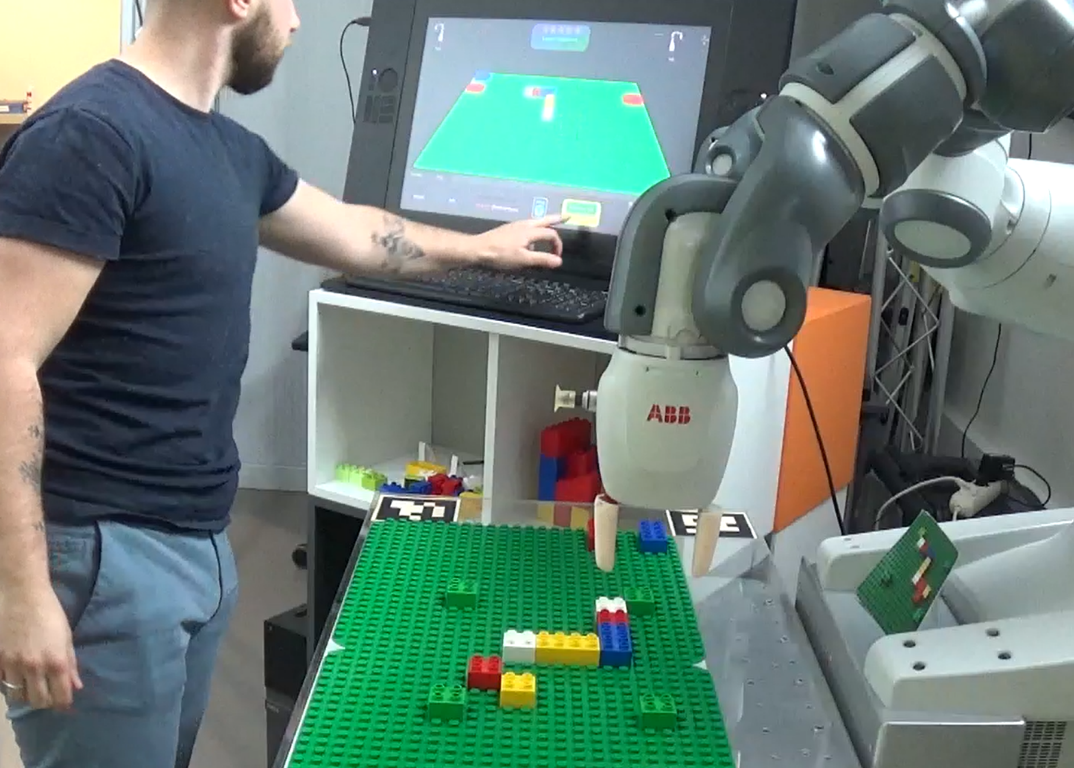}

\caption{A participant realizing the task with YuMi (H/C
condition)}
\end{figure}
In both conditions, they had to reproduce a model in front of them (A in
Figure 2 and 3) in a designated area (B) with Duplos located in an area
in front of them (C). In the H/H condition, they could talk to each
other. In the H/C condition, they had to interact using in interface
next to them (D). During the simple task, participants reproduced a
one-level model in collaboration as fast as possible (cf. Figure 3).

\begin{figure}
\centering\includegraphics[width=4.2003in,height=2.72373in]{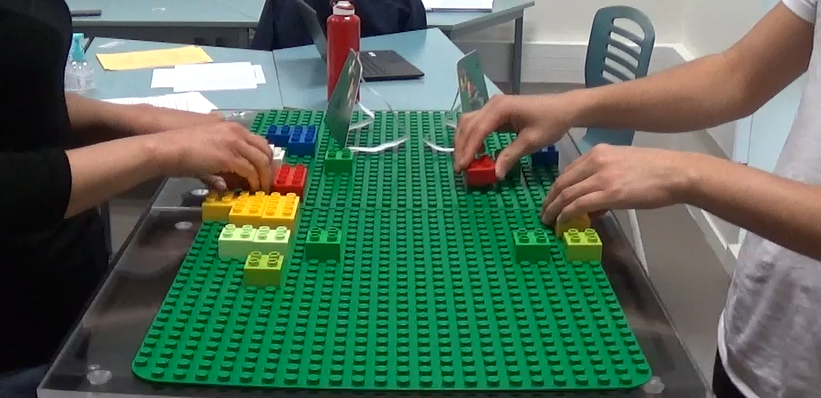}

\caption{Two participants realizing the task (H/H condition)}
\end{figure}

During the complex task, participants reproduced a 5-level model in
collaboration as fast as possible while answering simple math additions.
In both conditions, the participant had access to ¾ of the Duplo's stock
to realize the task. The human collaborator or the cobot had ¼ of the
Duplo's stock.

The robot used is a Yumi from ABB Robotics. Through a vision camera
system the cobot can automatically see what the human chooses to do.
This ability was programmed using Python language and OpenCV computer
vision library. After a perception stage, the system uses a
decision-making interface that uses a planning. The cobot decides what
the best next action is, taking into consideration its previous actions
and the human previous actions. This stage is performed using Automated
Planning techniques. Then, the selected action is executed by YuMi using
the RAPID language by ABB. This action is projected on a Graphical User
Interface to demonstrate the action to the participant.

After completing the assembly task, two questionnaires were filled by
participants. First, a translated version (Cegarra \& Morgado, 2009) of
the NASA-TLX (Hart \& Staveland, 1988) was used to assess perceived
workload (Table 1). This scale measures workload using subscales of the
mental load, physical load, temporal demand, frustration, effort and
performance (Hart, 2006). Participants passed the latest version of
NASA-TLX (see Cegarra \& Morgado, 2009) after each task (simple and
complex). The participant must answer on a 100-point Likert scale to
what degree he agrees with each sentence.

\begin{table}
\begin{center}
\caption{The NASA-TLX (Hart \& Staveland, 1988).}
\includegraphics[scale=0.4]{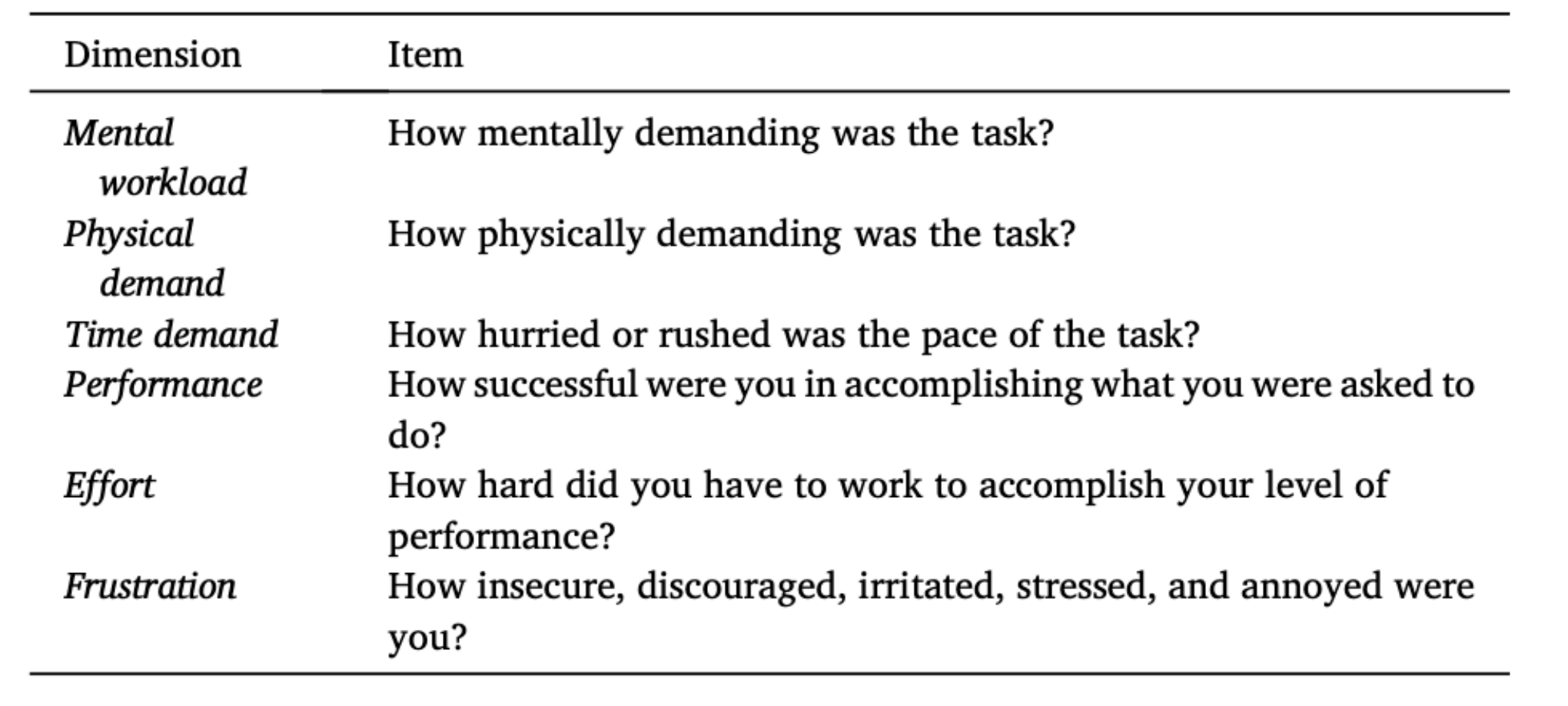}
\end{center}
\end{table}

An adapted version of the UTAUT2 questionnaire was used to assess the
acceptability of a collaboration with a cobot (Table 2). Martin (2018)
translated this questionnaire in French and adapted it by adding the
measure of ease of use developed by Davis (1989) and the measure of
trust developed by Martin (2018) to the original questionnaire of the
UTAUT2 (Venkatesh et al., 2012). The model fit of the questionnaire was
satisfactory and the items explained 89\% of the intent of use's
variance. In our version, ``this system'' was replaced by ``the cobot'',
and assessed the perceived ease of use, the perceived coherence for the
task, the perceived pleasure, the perceived usefulness, the perceived
social influence and the trust (Cronbach's alphas \textgreater{} 0,74).
Participants were asked with what degree they agree with the sentence
with a Likert scale from 1 (do not agree at all) to 7 (do totally
agree). A Principal Component Analysis (SPSS Version 28) revealed that 6
factors were measured by the 6 different scales (this is what we
expected) and that these factors combined explained 78,863\% of the
total variance of answers.

\begin{table}[!]
  \begin{center}
\caption{The acceptability questionnaire used in the experiment
(the questions were asked in French, they were only translated for the
article)}
\includegraphics[scale=0.5]{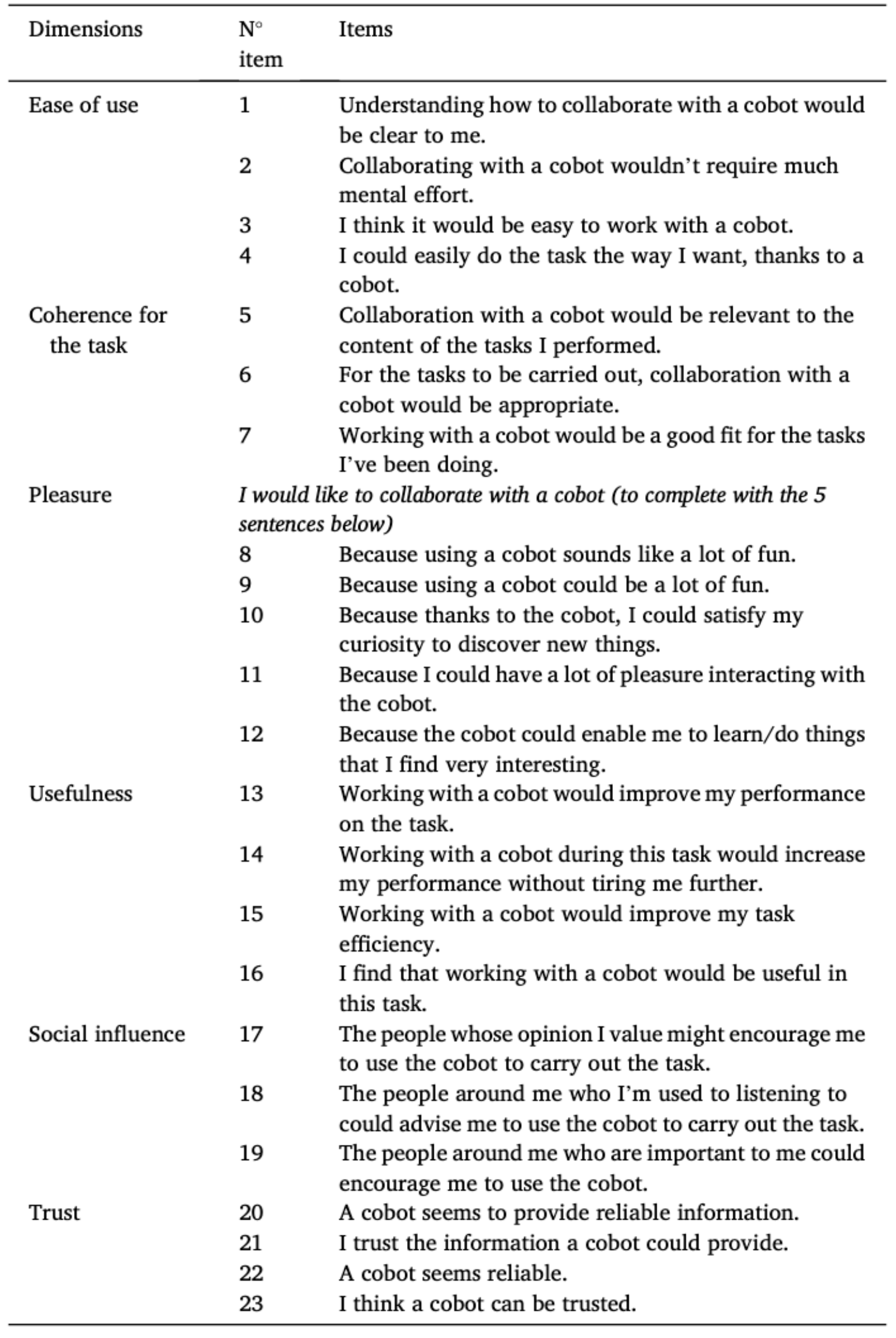}
\end{center}
\end{table}

The questionnaire was implemented using LimeSurvey software. This
software was made available thanks to SCREEN (Common Resource Service
for Experimentation and Digital Equipment) managed by the University of
Grenoble Alpes.

Participants were filmed during the tasks. These videos were used to
determine the number of gestures (number of arm movements of the
participant use to click on a button, pick and place a block, and
pointing), the number of errors (a block placed in the wrong spot) and
the time completion (from the ``start'' of the experimenter to the
``finished'' of the participant''). A task was considered ``successful''
when the model was correctly replicated (no misplaced or missing
blocks). The analysis of the videos was made with the software Boris
(Friard \& Gamba, 2016).

\section{Results}

Bugs (programming issues or lack of perception by the cobot) impacted
the experiments for 71\% of the H/C duos. Those issues were fully part
of the experiment as cobots are still in development, so the data
collected were not removed in case of bugs. It was considered to be a
part of the cobotic collaboration variable. On all the variables tested,
two participants were outliers on multiples variables (they were not
able to realize the task, even alone, due to spatial localization
problem). We decided to not include them in the statistical tests. The
final sample size for the study is 120 participants. Study results were
calculated using IBM SPSS 28. Multi-analysis of variance (MANOVA),
ANOVAs and mean comparisons were used. To calculate interaction effects,
generalized mixed model were used. Cohen's D was also calculated for the
effect sizes.

\subsection{Questionnaires validity}

Internal consistency of the NASA-TLX was confirmed: no item was
correlated more than 0.70 with another. The Cronbach's alpha was
satisfying (0.761). For the acceptability questionnaire, we performed a
principal component analysis to test the reliability of our
questionnaire and the validity of the different scales. We removed 4
problematic items (which are not in the right dimension or too close to
.50). By removing these 4 items, we explain almost 82\% of the
responses' variance. The questionnaire's Cronbach's alpha is 0.919.

\subsection{The moderative effect of the \textbf{task demand}
(Hypothesis 1)}

The table below presents the descriptive results of the first hypothesis
(cf. Figure 3).

\begin{figure}[!t]
  \centering
\caption{Bar chart of the workload, success, errors, time
completion and gestures in the simple and complex tasks during the
collaboration human/human compared to human/cobot}
\includegraphics[scale=0.4]{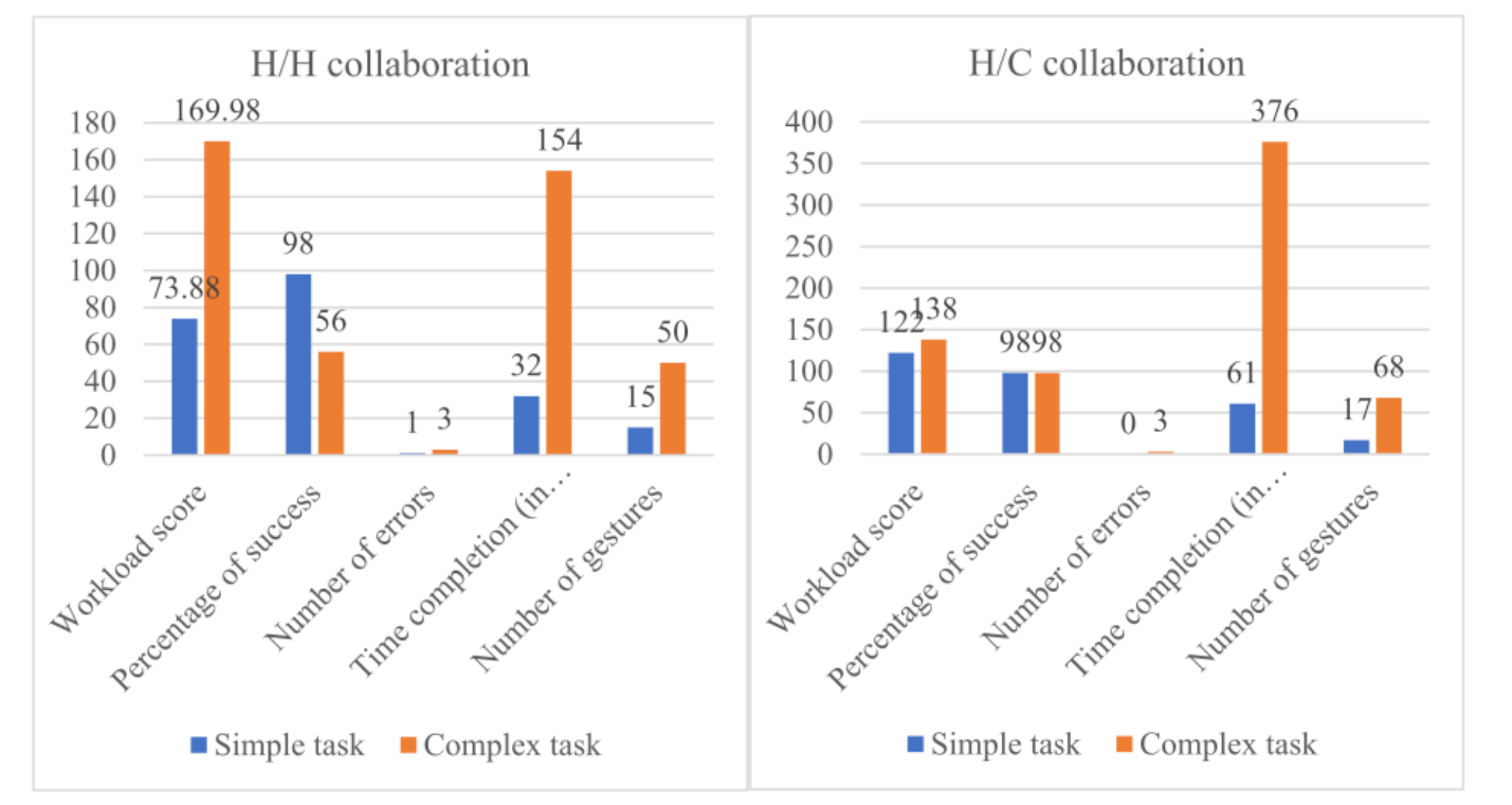}
\end{figure}

An interaction effect was found between task demand and workload
{[}\emph{f}(1)=61,134; \emph{p}\textgreater0,000{]}. The type of
collaboration moderates the effect of the type of task in terms of
workload. The effect of the task demand on the workload is lower in the
H/C condition than in the H/H condition (cf. Figure 4).

\begin{figure}[!t]
  \centering
\caption{Interaction effect of the type of collaboration on the
effect of task complexity on workload}
\includegraphics[scale=0.4]{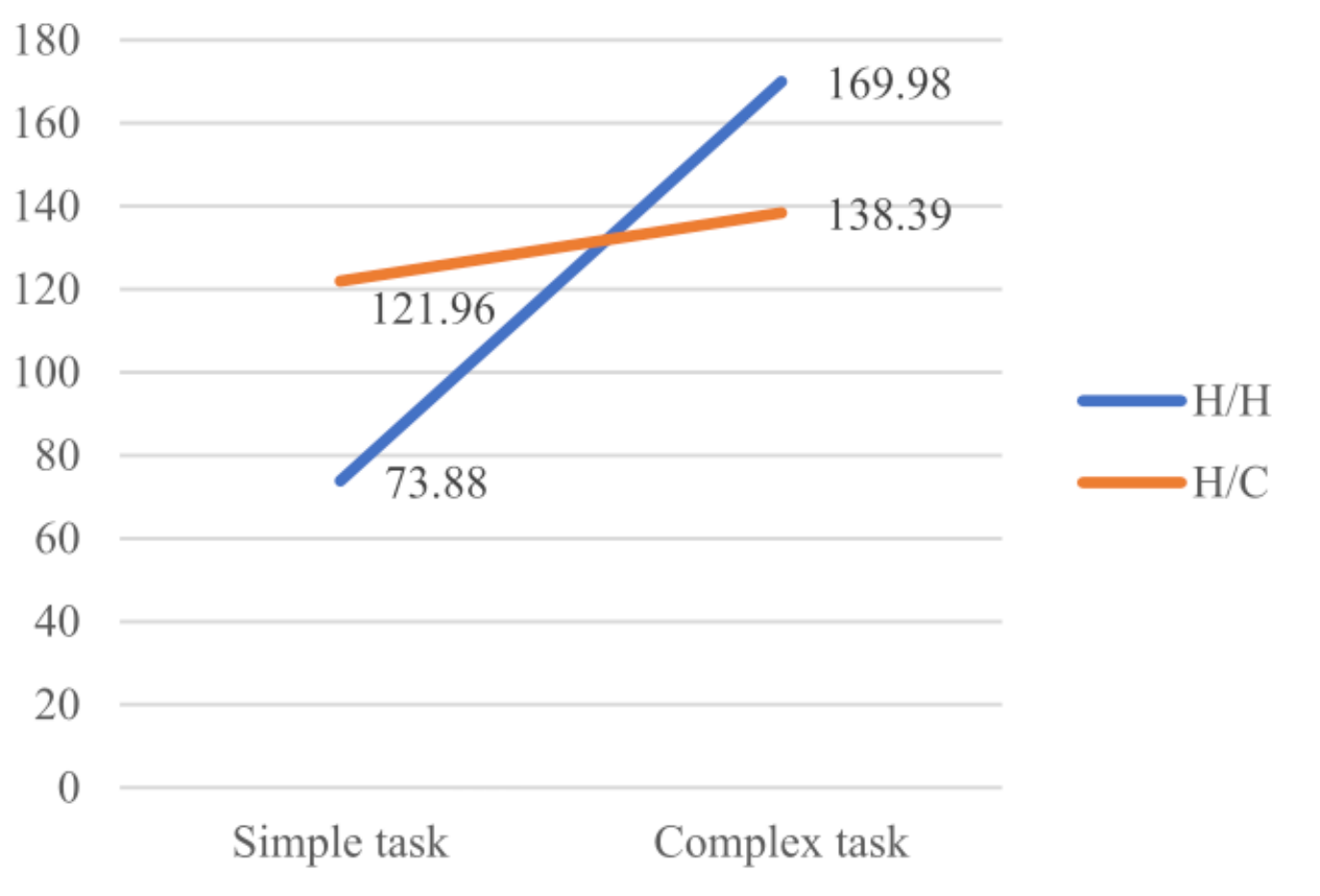}
\end{figure}

An interaction effect was found between task demand and success rate
{[}\emph{f}(1)=2,482; \emph{p}\textgreater0,000{]}. The type of
collaboration moderates the effect of the task demand in terms of
success rate. The effect of the task demand on the success is lower in
the H/C condition than in the H/H condition (cf. Figure 5).

\begin{figure}[!t]
  \centering
\caption{Interaction effect of the type of collaboration on the
effect of task complexity on the success percentage}
\includegraphics[scale=0.4]{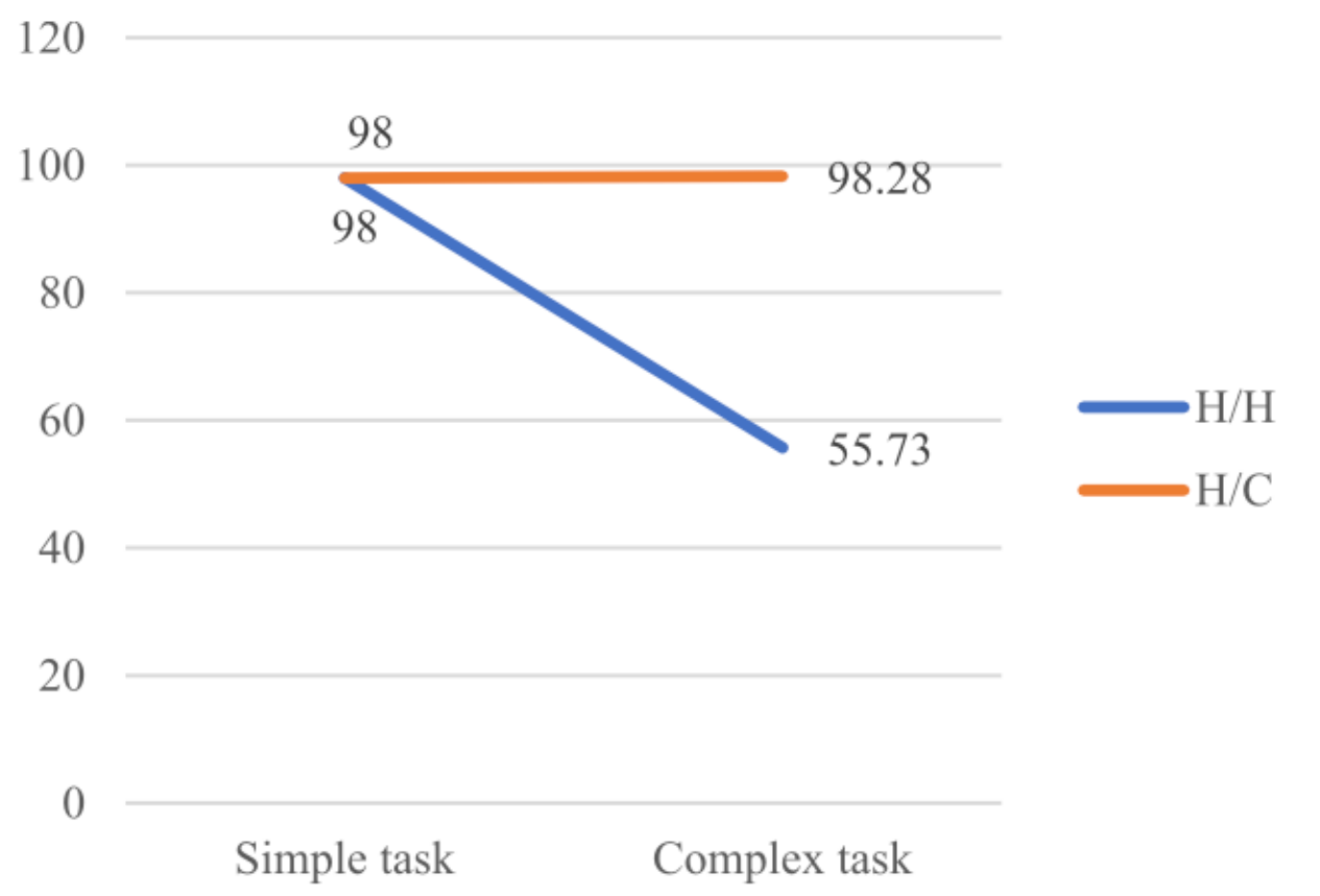}
\end{figure}

No interaction effect was found between task demand and number of errors
{[}\emph{f}(1)=1,088; p\textgreater0,299{]}.

An interaction effect was found between task demand and time completion.
The type of collaboration moderates the effect of the task demand in
terms of time completion {[}\emph{f}(1)=158,250; p\textgreater0,000{]}.
The effect of the task demand on the time completion is higher in the
H/C condition than in the H/H condition (cf. Figure 6).

\begin{figure}[!t]
  \centering
\caption{Interaction effect of the type of collaboration on the
effect of task complexity on time completion}
\includegraphics[scale=0.4]{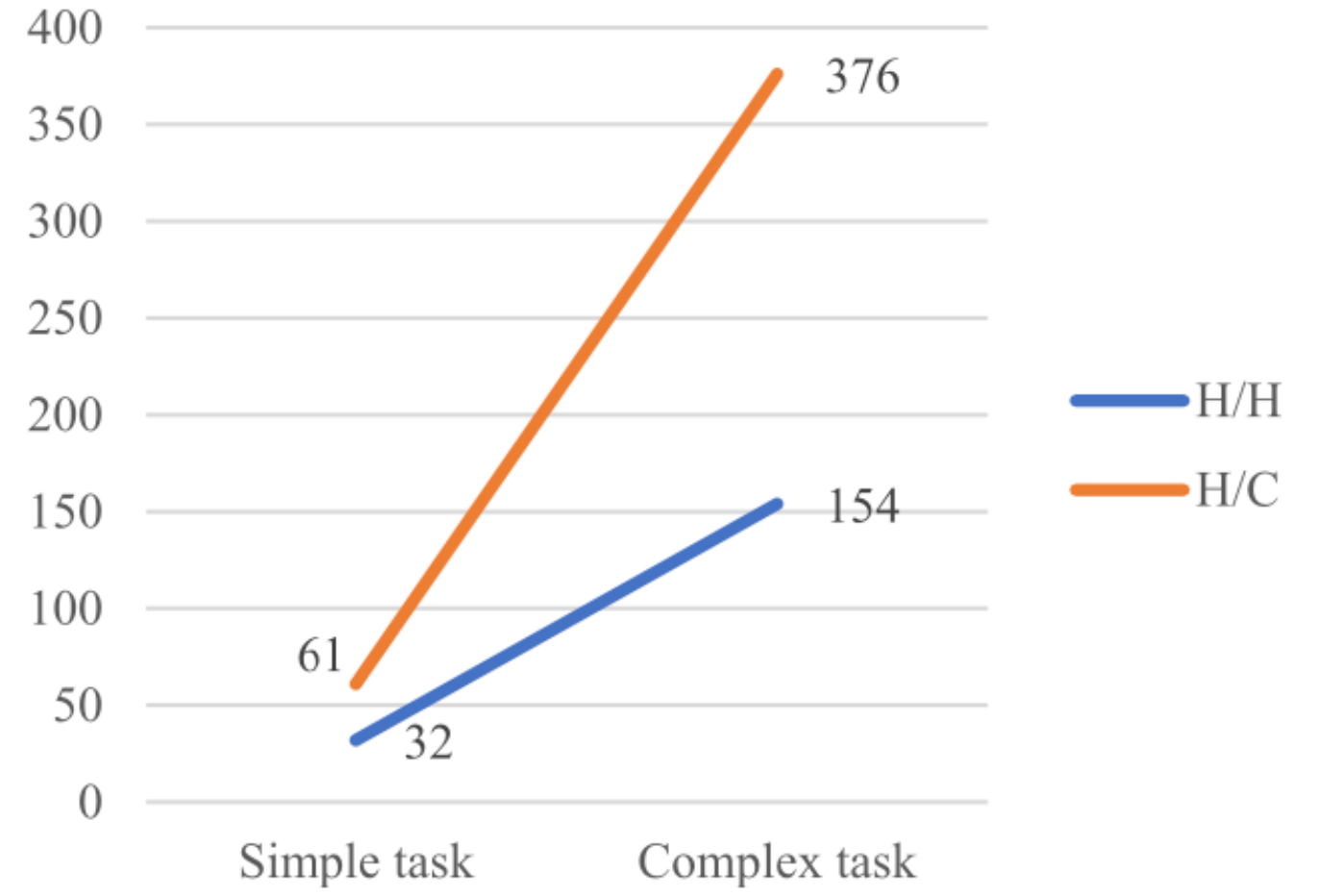}
\end{figure}

An interaction effect was found between task demand and number of
gestures. The type of collaboration moderates the effect of the task
demand in terms of number of gestures {[}\emph{f}(1)=29,066;
p\textgreater0,000{]}. The effect of the task demand on the number of
gestures is higher in the H/C condition than in the H/H condition (cf.
Figure 7).

\begin{figure}[!t]
  \centering
\caption{Interaction effect of the type of collaboration on the
effect of task complexity on number of gestures}
\includegraphics[scale=0.4]{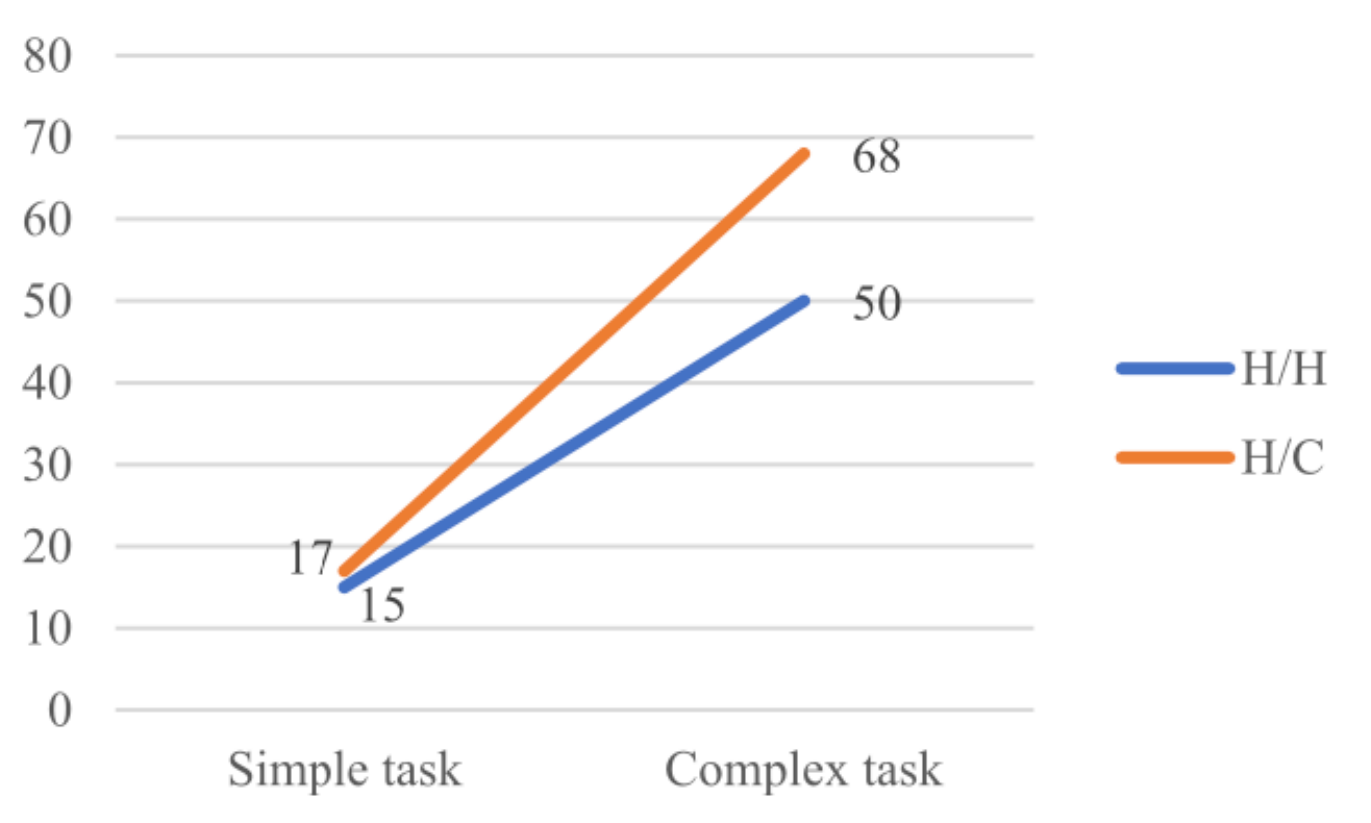}
\end{figure}

\subsection{H2: The moderation effect of cobot's adaptation to the
type dominant hand}

The effect of the cobot's adaptation to the type of dominant hand has no
effect on the different independent variables {[}\emph{f}(5)= 1,229;
\emph{p}=0,311{]}.

\subsection{H3: The effect of the type of collaboration}

The figure below presents the descriptive results of the third
hypothesis (Figure 8).

\begin{figure}[!t]
  \centering
\caption{Bar chart of the workload, success, errors, time
completion and acceptation in collaboration with a cobot and with
another human}
\includegraphics[scale=0.4]{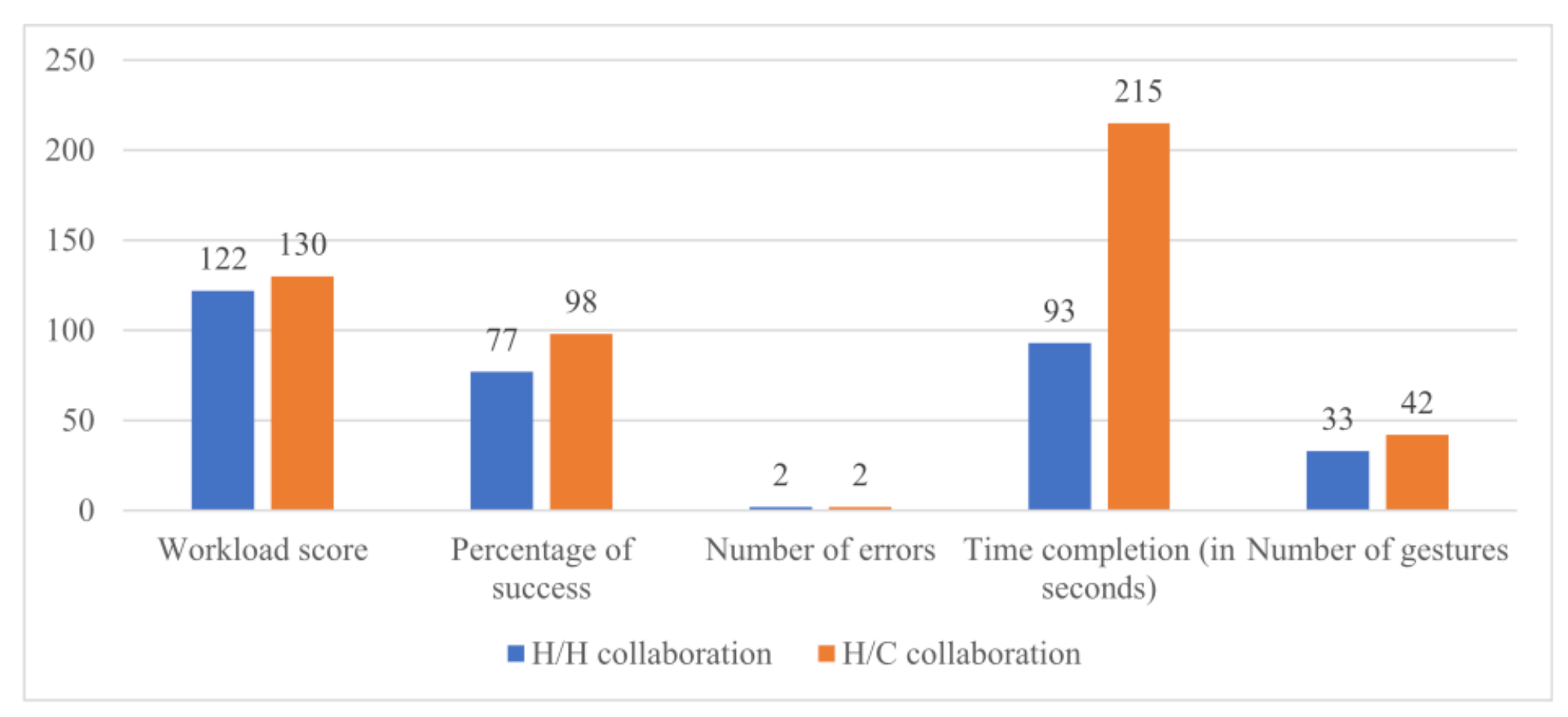}
\end{figure}

The type of collaboration has an effect on the success of the task, time
completion and number of gestures. However, it has no effect on workload
and number of errors (cf. Table 3).

\begin{sidewaystable}
%\begin{table}[!t]
  \centering
\caption{Results of the multi-analysis of variance of the type of
collaboration on workload, success, errors, time and gestures
(significant effects in bold)}
\includegraphics[scale=0.42]{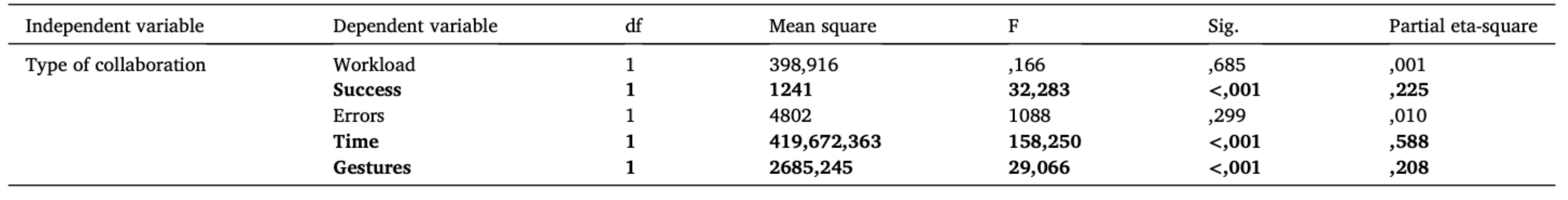}
%\end{table}
\end{sidewaystable}

The type of collaboration doesn't affect the workload globally. Also,
the workload after the simple task in the H/C condition is higher than
in the H/H condition {[}\emph{u}=2542,000; \emph{z}=4,327;
p\textless0,000{]}. The Cohen's D is -0,827, the effect is large.
However, the workload after the complex task in the H/C condition is
lower than in the H/H condition {[}\emph{u}=1338,500; \emph{z}=-2,154;
\emph{p}\textless0,000{]}. The Cohen's D is 0,464, the effect is
moderate.

There is no significant effect of the collaboration on the number
errors. However, participants did less errors by time spent on the task
in the H/C condition than in the H/H condition {[}\emph{t}(111)=5,988;
\emph{p}\textless0,001{]}. The Cohen's D is 1,130, the effect is large.

Also, bugs of the cobot impacted positively the time spent on the task
{[}\emph{t}(50)=1,966; \emph{p}=0,027{]} and the number of gestures
{[}\emph{u}=186,500; \emph{z}=-2,014; \emph{p}=0,044{]}. Both Cohen's D
indicate a moderate effect. The frequency of gesture (number of gesture
/ time completion) was lower with a cobot {[}\emph{t}(111)=-5,391;
\emph{p}\textless0,001{]}. Participants did less gestures considering
the time spent in the H/C condition than in the H/H condition. The
Cohen's D is 2,660, the effect is large.

\subsection{H4: The cobotic collaboration acceptability score}

The figure below presents the descriptive results of the third
hypothesis (cf. Figure 9).

\begin{figure}[!t]
  \centering
\caption{Bar chart of the subdimension of the acceptability
scale}
\includegraphics[scale=0.4]{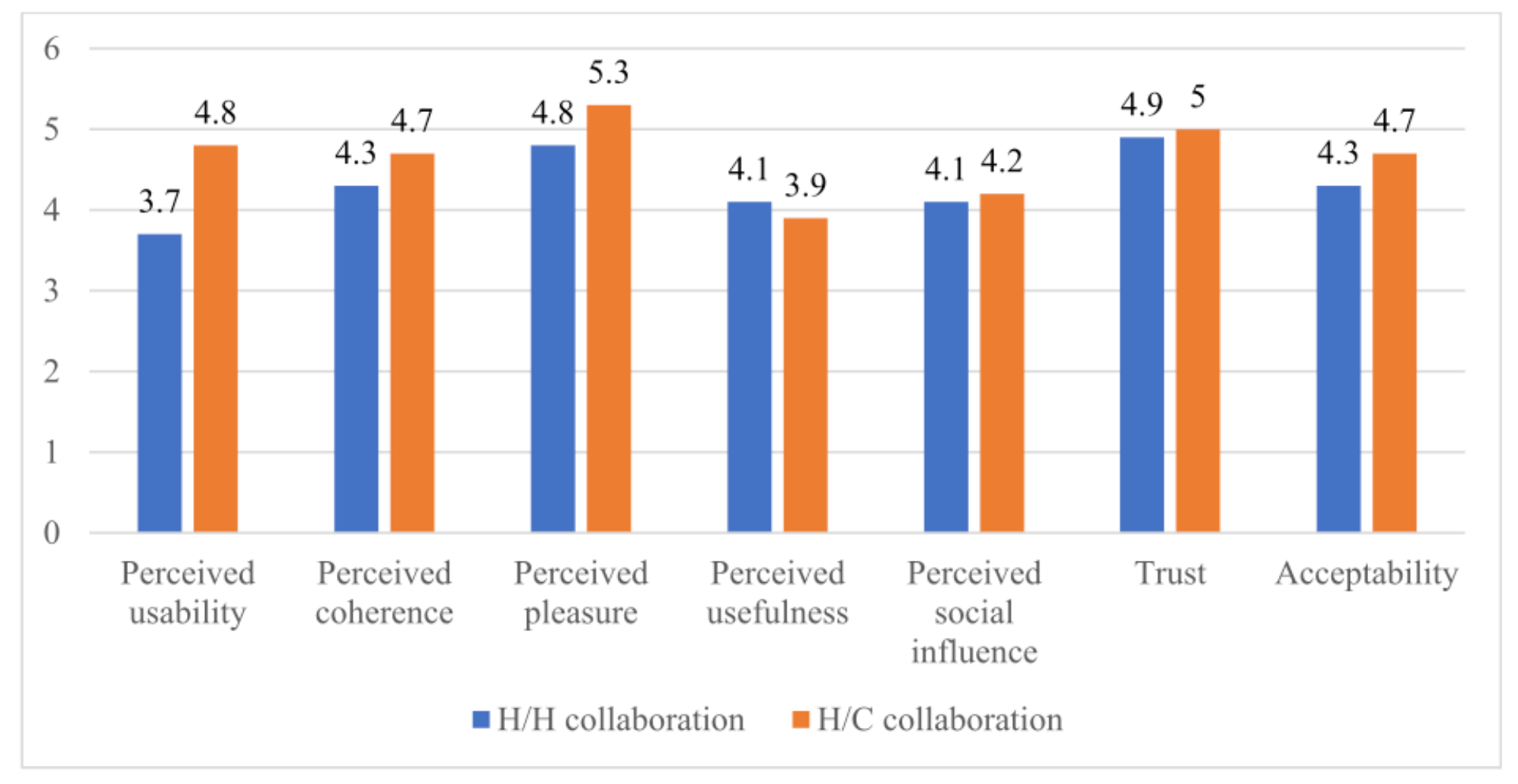}
\end{figure}

Working with a cobot has a positive effect on ease of use perceived and
on pleasure perceived (cf. Table 4).

\begin{sidewaystable}
%\begin{table}[!t]
  \centering
\caption{Results of the multi-analysis of variance of the type of
collaboration on the different scales of the acceptability questionnaire
(significant effects in bold)}
\includegraphics[scale=0.42]{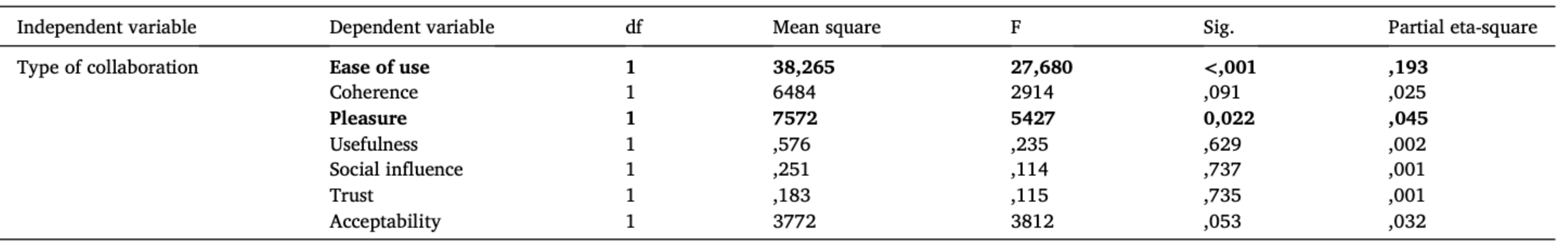}
%\end{table}
\end{sidewaystable}

\section{Discussion}

The objective of this study was to identify the impact of a H/C
collaboration in an industry-like assembly task. It has brought to light
some interesting and innovative results.

First, the cobotic collaboration has decreased the task demand's effect
on the human workload and increase the success of the task. During the
cobotic collaboration, the human's workload did not change
significantly. The success rate remained high no matter the level of
difficulty of the task. It was not the case during the collaboration
H/H. It led to a higher workload during a complex task and a lower
success rate. Also, whether the robot adapted to the dominant hand or
not, the results were no different. This could mean that this type of
adaptation is not necessary. Liu et al. (2022) found different results.
In their study, H/C collaboration did not have the same impact on the
workload and the performance according to the type of task they
realized. In our case, the difficulty changed, not the type of task.
Future studies should investigate both.

Second, H/C collaboration increased the effect of the task demand on the
time completion and the number of gestures. A higher time completion can
be potentially explained by the correction of errors. Indeed, in the H/C
condition, the success rate is significantly higher than in the H/H
condition but the number of errors is the same. This means that the
cobot encouraged participants to correct their errors, which involved
more gestures and a longer time completion. Furthermore, a test revealed
that the cobot's bugs had a positive impact on the number of gestures on
the time completion. The participants that collaborated with the cobot
experienced bugs that would need to be fixed by doing even more gestures
and by waiting for the bug to be fixed. Another study has compared H/H
and H/C collaboration and it showed no increase in the completion time
(Fager et al., 2019). However, the frequency of gestures is lower in the
H/C condition. Participants made fewer gestures with the cobot in
relation to the time spent on the task. This is coherent with a previous
experiment (Fournier et al., 2022).

The type of collaborations did not impact the workload as a whole, which
is coherent with other previous experiment (Fournier et al., 2022).
However, it increased the workload during the simple task and decreased
it during the complex task. We suspect a novelty effect (Elston, 2021).
This mixed effect would create bias in the workload scores. The fact
that the workload decreased during the complex task could mean that
working with a cobot is less demanding than working with another human
in terms of workload.

The type of collaboration had an effect on the success rate. In the H/C
condition, participants had a higher success rate than in the H/H
condition. This result is not surprising as others studies have found a
lower number of errors (Fournier et al., 2022) and a better output
quality (Salunkhe et al., 2019) in H/C collaborations. In our study, the
number of errors was equivalent in both conditions. However, the
frequency of errors was lower in the H/C condition.

Our study explored the H/C collaboration acceptability with 6 different
scales. Participants in the H/H condition had high perceptions of
pleasure and trust of H/C collaboration. This is coherent with another
study that found high scores in trust and pleasure perceived
(Cippelletti et al., in revision) in the working population.
Participants that worked with the cobot had high perceptions of ease of
use, coherence for the task, pleasure and trust of H/C collaboration.
There was a significative effect of the type of collaboration on the
ease of use and the pleasure perceived. H/C collaboration is perceived
as more pleasant and feasible when the participant tried this
collaboration before. As far as we know, those results have never been
found before.

This study has few limits: the experiments were impacted by bugs that
affected some results. Indeed, bugs increased the number of gestures
(probably because the participant had to replace some blocks) and the
time completion (the participant had to wait for the issue to be
resolved). However, all the other variables, including the workload,
were not impacted by the bugs. Participants may have been particularly
patient with the cobot, in a real industry setting, it might have been
different. In fact, the experiment has to be replicated with a better
performing cobot. It should also be replicated with more left-handed
participants even if they are less numerous in the population. We did
not take age into account in our study since we consider that this
variable was not useful to our analysis, but it could be one of the
variabilities to be considered. So, we will ask for the age in our next
study.

Also, the results bring some new research opportunities. It would be
interesting to test the acceptability in the work field with operators
that are used to work with cobots (Compan et al., 2023). This way, we
would be able to assess the situated acceptation, which is the
technology acceptance of concerned operators (Bobillier Chaumon, 2016,
2021). Also, testing the H/C collaboration impacts in different work
context would be beneficial for the industry.

Those results have practical implications: cobots are able to adapt to
human strategy in an industry-like task. It could decrease the
operator's workload and increase the quality of the results. However, we
discovered that the cobot adaption to the dominant hand is probably not
useful as it doesn't have an impact on the operator and on the results.
Also, working with a cobot could increase the time spent on the task.
Work engineers should address what is there main goal before
implementing a cobot (speed or quality of output).

\section{Conclusion}

This study has demonstrated that cobotic collaboration can balance the
effects of individual variability (strategy) and of task variability
(difficulty) on the output quality (success of the task) and on the
human workload. By doing so, it increased the time and the number of
gestures needed to realize the task (while decreasing their frequency).
In this experiment, the cobotic collaboration has proven to increase the
output quality and has reduced to human workload in a complicated task.
Also, it is the first study to compare the acceptability of cobotic
collaboration of people that have never experienced it with people that
just did. Results show that all participants have a high level of
pleasure perceived and trust associated with cobots. Having worked with
a cobot increases the ease of use and the pleasure perceived. Those
results can be of use to the industry that wants to implement cobots.

\paragraph{Conflict of interest}: the authors declare that no conflicts of
interest might have influence the research.

\paragraph{Funding}: this study was founded by the National Research Agency
as part of the AAPG 2020 (PACBOT project).

\section{References}

Alarcon, G. M., Gibson, A. M., Jessup, S. A., \& Capiola, A. (2021). Exploring the differential effects of trust violations in human-human and human-robot interactions. \emph{Applied Ergonomics}, \emph{93}, 103350. \url{https://doi.org/10.1016/j.apergo.2020.}

Barcellini, F., Béarée, R., Benchekroun, T.-H., Bounouar, M., Buchmann, W., Dubey, G., Lafeuillade, A.-C., Moricot, C., Rosselin-Bareille, C., Saraceno, M., \& Siadat, A. (2023). Promises of industry 4.0 under the magnifying glass of interdisciplinarity: Revealing operators and managers work and challenging collaborative robot design. \emph{Cognition, Technology \& Work}, \emph{25}(2‑3), 251‑271. https://doi.org/10.1007/s10111-023-00726-6

Bevan, N., Carter, J., \& Harker, S. (2015). ISO 9241-11 Revised: What
Have We Learnt About Usability Since 1998? In M. Kurosu (Ed.),
\emph{Human-Computer Interaction: Design and Evaluation} (pp. 143‑151).
Springer International Publishing. \url{
https://doi.org/10.1007/978-3-319-20901-2\_13}

Bobillier Chaumon, M.-E. (2016). L'acceptation située des technologies
dans et par l'activité\,: Premiers étayages pour une clinique de
l'usage. \emph{Psychologie du Travail et des Organisations},
\emph{22}(1), 4‑21. \url{https://doi.org/10.1016/j.pto.2016.01.001}

Bobillier Chaumon, M.-E. (2021). Exploring the Situated Acceptance of
Emerging Technologies in and Concerning Activity: Approaches and
Processes. In M.-E. Bobillier Chaumon (Ed.). \emph{Digital
Transformations in the Challenge of Activity and Work} (pp. 237‑256).
John Wiley \& Sons, Ltd. \url{https://doi.org/10.1002/9781119808343.ch18}

Bogue, R. (2016). Europe continues to lead the way in the collaborative
robot business. \emph{Industrial Robot: An International Journal},
\emph{43}(1), 6‑11. \url{https://doi.org/10.1108/IR-10-2015-0195}

Bouillet, K., Lemonnier, S., Clanche, F., \& Gauchard, G. (2023). Does
the introduction of a cobot change the productivity and posture of the
operators in a collaborative task? \emph{PLOS ONE}, \emph{18}(8),
e0289787. \url{https://doi.org/10.1371/journal.pone.0289787}

Brangier, E., \& Valléry, G. (2021). \emph{Ergonomie\,: 150 notions
clés}. Dunod.

Catchpole, K., Bisantz, A., Hallbeck, M. S., Weigl, M., Randell, R.,
Kossack, M., \& Anger, J. T. (2019). Human factors in robotic assisted
surgery: Lessons from studies `in the Wild'. \emph{Applied Ergonomics},
\emph{78}, 270‑276. \url{https://doi.org/10.1016/j.apergo.2018.02.011}

Cegarra, J., \& Morgado, N. (2009). \emph{Étude des propriétés de la
version francophone du NASA-TLX}. COOP'2000.
\url{https://docplayer.fr/21755956-Etude-des-proprietes-de-la-version-francophone-du-nasa-tlx.html}

Cheon, E., Schneiders, E., \& Skov, M. B. (2022). Working with Bounded
Collaboration: A Qualitative Study on How Collaboration is
Co-constructed around Collaborative Robots in Industry.
\emph{Proceedings of the ACM on Human-Computer Interaction},
\emph{6}(CSCW2), 369:1-369:34. \url{https://doi.org/10.1145/3555094}

Cippelletti, E. (2017). \emph{Aide à la conception, test de l'usage et
de l'acceptation d'un logiciel de maintenance} {[}These de doctorat,
Université Grenoble Alpes (ComUE){]}. \url{https://www.theses.fr/2017GREAH038}

Colgate, J. E., Wannasuphoprasit, W., \& Peshkin, M. A. (1996).
\emph{Cobots: Proceedings of the 1996 ASME International Mechanical
Engineering Congress and Exposition}, 433‑439.
\url{http://www.scopus.com/inward/record.url?scp=0030402971\&partnerID=8YFLogxK}

Compan, N., Matias, J., Lutz, M., Rix-Lièvre, G., Brissaud, D.,
Belletier, C., \& Coutarel, F. (2023). The impact of Enabling
Collaborative Situations in AR-assisted consignment tasks.
\emph{Frontiers in Industrial Engineering}, \emph{1}, 1258241.
\url{https://doi.org/10.3389/fieng.2023.1258241}

Davis, F. D. (1989). Perceived Usefulness, Perceived Ease of Use, and
User Acceptance of Information Technology. \emph{MIS Quarterly},
\emph{13}(3), Article 3. \url{https://doi.org/10.2307/249008}

Djuric, A., Rickli, J., Sefcovic, J., Hutchison, D., \& Goldin, M. M.
(2019, janvier 15). \emph{Integrating Collaborative Robots in
Engineering and Engineering Technology Programs}. ASME 2018
International Mechanical Engineering Congress and Exposition.
\url{https://doi.org/10.1115/IMECE2018-88147}

Elston, D. M. (2021). The novelty effect. \emph{Journal of the American
Academy of Dermatology}, \emph{85}(3), 565‑566.
\url{https://doi.org/10.1016/j.jaad.2021.06.846}

European Commission. (2021). \emph{Industry 5.0: Towards a sustainable,
human centric and resilient European industry} (KI-BD-20-021-FR-N).
Publications Office of the European Union.
\url{https://data.europa.eu/doi/10.2777/308407}

Fager, P., Calzavara, M., \& Sgarbossa, F. (2019). Kit Preparation with
Cobot-supported Sorting in Mixed Model Assembly.
\emph{IFAC-PapersOnLine}, \emph{52}(13), 1878‑1883.
\url{https://doi.org/10.1016/j.ifacol.2019.11.476}

Fournier, É., Kilgus, D., Landry, A., Hmedan, B., Pellier, D., Fiorino,
H., \& Jeoffrion, C. (2022). The Impacts of Human-Cobot Collaboration on
Perceived Cognitive Load and Usability during an Industrial Task: An
Exploratory Experiment. \emph{IISE Transactions on Occupational
Ergonomics and Human Factors}, \emph{10}(2), 1‑8.
\url{https://doi.org/10.1080/24725838.2022.2072021}

Fruggiero, F., Lambiase, A., Panagou, S., \& Sabattini, L. (2020).
Cognitive Human Modeling in Collaborative Robotics. \emph{Procedia
Manufacturing}, \emph{51}, 584‑591.
\url{https://doi.org/10.1016/j.promfg.2020.10.082}

Gualtieri, L., Rauch, E., \& Vidoni, R. (2021). Emerging research fields
in safety and ergonomics in industrial collaborative robotics: A
systematic literature review. \emph{Robotics and Computer-Integrated
Manufacturing}, \emph{67}, 101998.
\url{https://doi.org/10.1016/j.rcim.2020.101998}

Guérin, F., Laville, A., Daniellou, F., Durrafourg, J., \& Kerguelen, A.
(1997). \emph{\hspace{0pt}Comprendre le travail pour le transformer}
(3ème édition). ANACT éditions.

Hart, S. G. (2006). \emph{NASA-task load index (NASA-TLX); 20 years
later}. \emph{50}, 904‑908.

Hart, S. G., \& Staveland, L. E. (1988). Development of NASA-TLX (Task
Load Index): Results of Empirical and Theoretical Research. In P. A.
Hancock \& N. Meshkati (Eds.), \emph{Advances in Psychology} (Vol. 52,
pp. 139‑183). North-Holland.
\url{https://doi.org/10.1016/S0166-4115(08)62386-9}

Hiatt, L. M., Harrison, A. M., \& Trafton, J. (2011).
\emph{Accommodating Human Variability in Human-Robot Teams through
Theory of Mind}.
\url{https://www.semanticscholar.org/paper/Accommodating-Human-Variability-in-Human-Robot-of-Hiatt-Harrison/e1682c374e701733b3b232f3d8f132697aa98992}

Hopko, S. K., Mehta, R. K., \& Pagilla, P. R. (2023). Physiological and
perceptual consequences of trust in collaborative robots: An empirical
investigation of human and robot factors. \emph{Applied Ergonomics},
\emph{106}, 103863. \url{https://doi.org/10.1016/j.apergo.2022.103863}

ISO. (2018). \emph{ISO 9241-11}. ISO.
\url{https://www.iso.org/obp/ui/\#iso:std:iso:9241:-11:ed-2:v1:en}

Kildal, J., Martín, M., Ipiña, I., \& Maurtua, I. (2019). Empowering
assembly workers with cognitive disabilities by working with
collaborative robots: A study to capture design requirements.
\emph{Procedia CIRP}, \emph{81}, 797‑802.
\url{https://doi.org/10.1016/j.procir.2019.03.202}

Kildal, J., Tellaeche, A., Fernández, I., \& Maurtua, I. (2018).
Potential users' key concerns and expectations for the adoption of
cobots. \emph{Procedia CIRP}, \emph{72}, 21‑26.
\url{https://doi.org/10.1016/j.procir.2018.03.104}

Knudsen, M., \& KaiVo-Oja, J. (2020). Collaborative Robots: Frontiers of
Current Literature. \emph{Journal of Intelligent Systems: Theory and
Applications}, 13‑20. \url{https://doi.org/10.38016/jista.682479}

Lesgauchers. (2017, octobre 31). \emph{9 millions de gauchers \ldots{}}.
\url{https://www.lesgauchers.com/information/gaucher-9-millions-en-france}

Liu, L., Schoen, A. J., Henrichs, C., Li, J., Mutlu, B., Zhang, Y., \&
Radwin, R. G. (2022). Human Robot Collaboration for Enhancing Work
Activities. \emph{Human Factors}, 00187208221077722.
\url{https://doi.org/10.1177/00187208221077722}

Mariscal, M. A., Ortiz Barcina, S., García Herrero, S., \& López Perea,
E. M. (2023). Working with collaborative robots and its influence on
levels of working stress. \emph{International Journal of Computer
Integrated Manufacturing}, 1‑20.
\url{https://doi.org/10.1080/0951192X.2023.2263428}

Martin, N. P. Y. (2018). \emph{Acceptabilité, acceptation et expérience
utilisateur\,: Évaluation et modélisation des facteurs d'adoption des
produits technologiques} {[}Phdthesis, Université Rennes 2{]}.
\url{https://tel.archives-ouvertes.fr/tel-01813563}

Matheson, E., Minto, R., Zampieri, E. G. G., Faccio, M., \& Rosati, G.
(2019). Human--Robot Collaboration in Manufacturing Applications: A
Review. \emph{Robotics}, \emph{8}(4), 100.
\url{https://doi.org/10.3390/robotics8040100}

Müller, R., Vette, M., \& Geenen, A. (2017). Skill-based Dynamic Task
Allocation in Human-Robot-Cooperation with the Example of Welding
Application. \emph{Procedia Manufacturing}, \emph{11}, 13‑21.
\url{https://doi.org/10.1016/j.promfg.2017.07.113}

Nielsen, J. (1994). \emph{Usability Engineering}. Morgan Kaufmann
Publishers Inc.

Oztemel, E., \& Gursev, S. (2020). Literature review of Industry 4.0 and
related technologies. \emph{Journal of Intelligent Manufacturing},
\emph{31}(1), 127‑182. \url{https://doi.org/10.1007/s10845-018-1433-8}

Pauliková, A., Gyurák Babeľová, Z., \& Ubárová, M. (2021). Analysis of
the Impact of Human--Cobot Collaborative Manufacturing Implementation on
the Occupational Health and Safety and the Quality Requirements.
\emph{International Journal of Environmental Research and Public
Health}, \emph{18}(4), 1927. \url{https://doi.org/10.3390/ijerph18041927}

Peshkin, M. A., \& Colgate, J. E. (1999). Cobots. \emph{Industrial
Robot}, \emph{26}(5), 335‑341.

Salunkhe, O., Stensöta, O., Åkerman, M., Berglund, Å. F., \& Alveflo,
P.-A. (2019). Assembly 4.0: Wheel Hub Nut Assembly Using a Cobot.
\emph{IFAC-PapersOnLine}, \emph{52}(13), 1632‑1637.
\url{https://doi.org/10.1016/j.ifacol.2019.11.434}

Schoose, C. (2022). \emph{From traditional deburring to cobot use:
Analysis of the professional gesture in order to prevent
musculosquelettal disorders} {[}Phdthesis, Université Grenoble Alpes
{[}2020-....{]}{]}. \url{https://theses.hal.science/tel-03942554}

Schoose, C., Cuny-Guerrier, A., Caroly, S., Claudon, L., Wild, P., \&
Savescu, A. (2023). Evolution of the biomechanical dimension of the
professional gestures of grinders when using a collaborative robot.
\emph{International Journal of Occupational Safety and Ergonomics},
\emph{29}(2), 668‑675. \url{https://doi.org/10.1080/10803548.2022.2065063}

Varrecchia, T., Chini, G., Tarbouriech, S., Navarro, B., Cherubini, A.,
Draicchio, F., \& Ranavolo, A. (2023). The assistance of BAZAR robot
promotes improved upper limb motor coordination in workers performing an
actual use-case manual material handling. \emph{Ergonomics}, 1‑18.
\url{https://doi.org/10.1080/00140139.2023.2172213}

Venkatesh, V., Morris, M. G., Davis, G. B., \& Davis, F. D. (2003). User
Acceptance of Information Technology: Toward a Unified View. \emph{MIS
Quarterly}, \emph{27}(3), Article 3. \url{https://doi.org/10.2307/30036540}

Venkatesh, V., Thong, J. Y. L., \& Xu, X. (2012). Consumer Acceptance
and Use of Information Technology: Extending the Unified Theory of
Acceptance and Use of Technology. \emph{MIS Quarterly}, \emph{36}(1),
Article 1. \url{https://doi.org/10.2307/41410412}

Verhulst, E. (2018). \emph{Contribution de l'étude de l'interaction en
environnement virtuel: Intérêt de la charge mentale} {[}Phdthesis,
Université d'Angers{]}. \url{https://tel.archives-ouvertes.fr/tel-02157566}

Xu, X., Lu, Y., Vogel-Heuser, B., \& Wang, L. (2021). Industry 4.0 and
Industry 5.0---Inception, conception and perception. \emph{Journal of
Manufacturing Systems}, \emph{61}, 530‑535.
\url{https://doi.org/10.1016/j.jmsy.2021.10.006}

\end{document}